\documentclass[sigconf,nonacm]{acmart}

\usepackage{amsmath}
\usepackage{algorithm}
\usepackage{algpseudocode}
\usepackage{booktabs}
\usepackage{makecell}
\usepackage{graphicx}
\usepackage{tabularx}
\usepackage{xcolor}
\renewcommand\footnotetextcopyrightpermission[1]{}
\newcommand{\system}{\textsc{Layerwise Repacking}}
\newcommand{\layoutF}{\ensuremath{\mathcal{F}}}
\newcommand{\layoutS}{\ensuremath{\mathcal{S}}}
\newcommand{\pp}{\,pp}

\title{\textsc{Layerwise Repacking}: Variable-Length Hybrid-Attention Training without an Extra Redistribution Pass}
\author{Kling Infra Team}
\ifdefined\correspondingauthor
  \correspondingauthor
\fi
\affiliation{%
  \institution{KlingAI Research}
  \city{Beijing}
  \country{China}
}

\author{Jiahao Zhang}
\authornote{This work was conducted during the authors' internship at KlingAI.}
\email{zhangjiahao09@kuaishou.com}
\affiliation{%
  \institution{KlingAI Research}
  \city{Beijing}
  \country{China}
}

\author{Junwei Feng}
\authornotemark[1]
\email{fengjunwei@kuaishou.com}
\affiliation{%
  \institution{KlingAI Research}
  \city{Beijing}
  \country{China}
}

\author{Dongliang Huang}
\authornotemark[1]
\email{huangdongliang@kuaishou.com}
\affiliation{%
  \institution{KlingAI Research}
  \city{Beijing}
  \country{China}
}

\author{Xiaoling Zhou}
\authornotemark[1]
\email{zhouxiaoling03@kuaishou.com}
\affiliation{%
  \institution{KlingAI Research}
  \city{Beijing}
  \country{China}
}

\author{Yiqiao Liao}
\authornotemark[2]
\email{liaoyiqiao@kling.ai}
\affiliation{%
  \institution{KlingAI Research}
  \city{Beijing}
  \country{China}
}

\author{Chengru Song}
\authornote{Corresponding authors.}
\email{songchengru@kuaishou.com}
\affiliation{%
  \institution{KlingAI Research}
  \city{Beijing}
  \country{China}
}

\begin{document}

\begin{abstract}

Variable-length training creates conflicting parallelism requirements across the layers of a hybrid-attention model. The longest samples require wide sequence-parallel (SP) groups of 16 or 32 ranks for full attention, whose cost grows quadratically with sequence length. Sparse-attention layers, dominated by near-linear MoE work, can instead balance most evaluated workloads within node-local SP8 groups. Reusing the wide groups for these layers pushes their all-to-all communication across nodes, where the observed effective rate is 7.0× lower. Yet switching to smaller groups changes which ranks own each sample; a separate redistribution at every full–sparse boundary would introduce another data transfer.

\system{} keeps each SP16/32 parent for full attention and repacks its
samples into node-local sparse children at SP$\{1,2,4,8\}$, where hidden
states persist between full layers.  The layout change rides on the two
existing full-layer Ulysses sites: at each, one variable-size all-to-all
performs both the head exchange and the sample repacking, so no
redistribution pass is added.  Because the balance the children can reach
depends on which samples share a parent, an auxiliary parent solver
rebalances parent membership before the children are formed.

On all 22 steps with SP16/32 parents among the first 200 training steps of
a 1F3S hybrid-attention model (one full-attention layer per three
sparse-attention layers), repacking alone reduces the mean end-to-end time
by 5.95\% and raises model FLOPs utilization (MFU) by 3.75 percentage
points over the baseline with repacking disabled; with the improved
solver, end-to-end time falls by 7.61\% and MFU rises by 4.83 percentage
points.
\end{abstract}

\maketitle

\section{Introduction}

\begin{figure*}[!t]
  \centering
  \includegraphics[width=\textwidth]{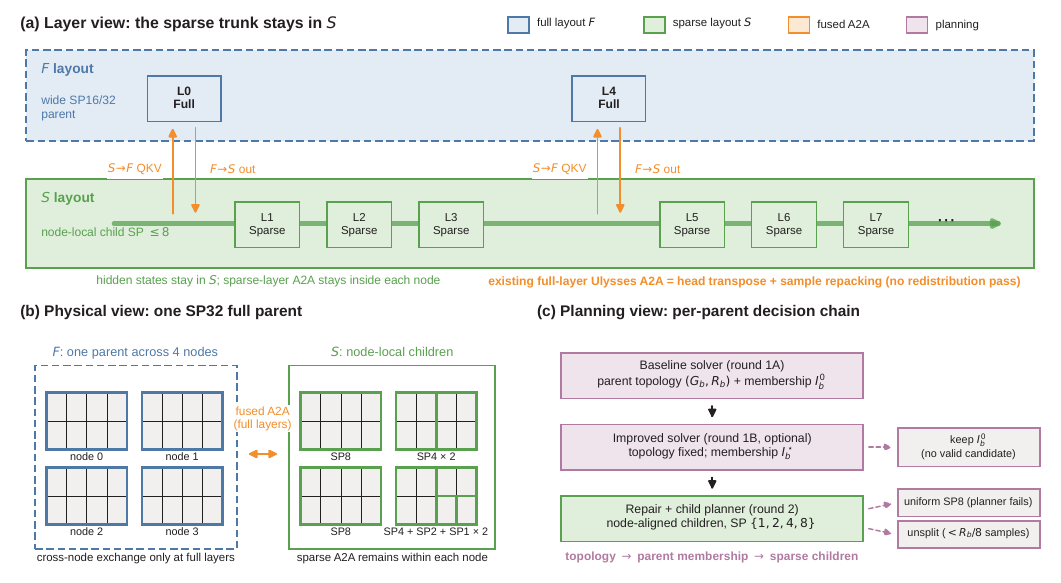}
  \caption{Overview of \system{}.
(a) Hidden states use the node-local sparse layout \layoutS{} between
full-attention layers; transitions to the wide layout \layoutF{} are
fused with the existing Ulysses exchanges.
(b) An SP32 full-attention parent spans four nodes, while its
sparse-layer children remain node-local.
(c) Planning assigns samples to parents, optionally rebalances parent
membership, and then forms sparse-layer children; dashed arrows
indicate fallback paths.}
  \Description{Three-panel diagram. The layer view shows hidden states staying
  in the node-local sparse layout and visiting the wide full layout only at
  full-attention layers. The physical view shows one SP32 parent over four
  nodes split into node-local children. The planning view shows rounds 1A, 1B,
  and 2 with their fallback paths.}
  \label{fig:overview}
\end{figure*}

Full attention captures long-range dependencies, but its quadratic cost makes it expensive at long sequence lengths. Hybrid-attention models therefore interleave a small number of full-attention layers with cheaper alternatives, retaining periodic global information flow while reducing training cost. Examples include Gemma~2's alternating local and global
layers~\cite{gemma2024gemma2}, MiniMax-01's one softmax layer after every
seven linear-attention layers~\cite{minimax2025minimax01}, Jamba’s interleaving of attention and Mamba layers~\cite{lieber2024jamba}, and Kimi Linear’s three linear-attention layers per full-attention layer~\cite{kimi2025linear}.
In the 1F3S pattern studied here, every full-attention layer is followed by
three sparse-attention layers. As these cheaper layers become more efficient, an increasing fraction of the model no longer needs the degree of parallelism required by full attention.

Variable-length training creates distinct parallelism requirements for
full-attention and sparse-attention layers. Here, each \emph{full} or
\emph{sparse} layer denotes an entire Transformer block, including
attention and MoE, classified by its attention type. In the evaluated
steps, the longest sample is $9.8$--$175\times$ the length of the shortest
(median: $23\times$). Because full-attention computation scales
quadratically with sequence length and its activations must be
distributed, the longest samples require sequence-parallel (SP) groups
of 16 or 32 ranks (SP16/32).
The sparse layers of the profiled
model, by contrast, are dominated by approximately length-linear MoE and
other per-token work, so their share of the same samples' work balances
within one eight-GPU node (Section~\ref{sec:gaps}).

Placement systems for variable-length training choose one group per sample
or packed micro-batch and use it for every
layer~\cite{wang2025flexsp,ge2025bytescale,li2026dynamiccp}.  A wide group
sized for the longest samples' full attention is then carried through the
three sparse layers that follow, moving their Ulysses
all-to-all~\cite{jacobs2023deepspeed} from the intra-node fabric to
inter-node links.  In our workload, this lowers the observed effective
Ulysses rate by $7.0\times$ relative to node-local SP8 groups, and
all-to-all occupies 21.4\% of a wide-parent step on wide ranks, against
4.3\% on SP8 ranks (Section~\ref{sec:mismatch}).

Assigning node-local groups to sparse layers requires more than changing
their sequence-parallel degree.  Because the full- and sparse-layer groups
place each sample's tokens on different ranks, every transition between
the two layer types requires an activation-layout conversion.  Implemented
as a separate redistribution phase, this conversion adds cross-node
activation traffic on top of the full layer's existing exchange.  In our
workloads, such a phase would offset approximately 30--60\% of the
communication savings from node-local sparse-layer execution, depending
on the data redistributed
(Section~\ref{sec:ownership}).

Accordingly, we propose \system{} to enable layer-specific parallel layouts without
adding a dedicated activation-redistribution pass, as shown in Figure~\ref{fig:overview},
where full-attention parents \layoutF{} keep SP$\{8,16,32\}$, while the samples of
each SP16/32 parent are repacked into node-local sparse children \layoutS{}
with SP$\{1,2,4,8\}$.  Hidden states persist in \layoutS{}, which we call the
sparse trunk, and visit \layoutF{} only inside full attention.  The key
observation is that each full layer's mandatory Ulysses exchange already
moves every token across the parent group: generalizing its two call sites
to variable split sizes and a deterministic permutation lets the same
all-to-all perform the standard sequence/head transpose and the sample
repacking between \layoutS{} and \layoutF{}, so no redistribution pass or
second payload transfer is added.

Child planning can redistribute work only within each parent, so its
achievable load balance depends on parent membership.  The baseline
solver (round~1A) assigns samples to parents using weights calibrated
to full-attention compute, as in schedulers for homogeneous Transformer
stacks~\cite{wang2025flexsp,ge2025bytescale,li2026dynamiccp}.  When full
and sparse layers use different parallel groups, these weights do not
capture sparse-layer costs and may place several expensive samples in
the same parent; child planning cannot redistribute them across parents.
We therefore introduce an auxiliary parent solver (round~1B) that
rebalances membership using separate full- and sparse-layer cost
components while preserving each parent's rank interval.  The child
planner (round~2) then partitions the revised parents
(Section~\ref{sec:planner}).

We evaluate \system{} on all 22 training steps with SP16/32 parents among
the first 200 steps of a 1F3S hybrid-attention model.  By separately
enabling repacking and the auxiliary parent solver, we isolate their
contributions: repacking alone reduces mean end-to-end time by 5.95\%,
while the complete system reduces it by 7.61\% and increases model FLOPs
utilization by 4.83 percentage points, relative to the baseline without
repacking.  These gains apply to steps that trigger wide parents rather
than the full 200-step window.  Repacking improves all 20 steps with a
single SP16 parent, but slows the two remaining steps, where an extremely
long sample limits the load balance achievable within node-local children.

Overall, our main contributions are as follows:
\begin{itemize}
  \item We propose \system{}, which assigns full- and sparse-attention
  layers different sequence-parallel layouts and fuses layout conversion
  into the existing Ulysses exchanges without an additional
  redistribution pass.

  \item We design an auxiliary parent solver that accounts for both
  full- and sparse-layer costs when assigning samples to parent groups,
  improving the load balance achievable by node-local children.

  \item We evaluate the framework on variable-length 1F3S training
  workloads, quantify the individual and combined gains of repacking
  and parent rebalancing, and identify cases where node-local repacking
  does not improve performance.
\end{itemize}

\section{Motivation}
\label{sec:motivation}

We first characterize the variable-length workload
(Section~\ref{sec:gaps}) and show why carrying a full-attention group
into sparse layers makes cross-node communication a bottleneck
(Section~\ref{sec:mismatch}).  We then examine two constraints on
node-local sparse-layer execution: layout transitions change token
ownership (Section~\ref{sec:ownership}), and achievable child balance
depends on parent membership (Section~\ref{sec:composition}).
These observations motivate pure Ulysses as the communication substrate
(Section~\ref{sec:whyup}).

\subsection{Sample and Layer Heterogeneity}
\label{sec:gaps}

Two gaps shape the workload: an inter-sample gap between samples of one
step and an inter-layer gap between layer types for the same sample.
Between samples, the gap is visible directly in the sample lengths: on
the 22 wide-parent steps evaluated in Section~\ref{sec:evaluation}, the
longest sample of each step is $9.8$--$175\times$ (median $23\times$) as
long as the shortest.  Because full attention is quadratic in length, the
gap in full-layer cost is larger still.  Between layers, the gap is set by
the design of the cheaper layer: a full-attention query attends to all $L$
tokens of its sample, whereas sparse attention bounds the entries one
query attends to, and a linear-attention layer keeps a constant-size
state.  DeepSeek-V3.2's sparse attention selects 2048 tokens per query,
reducing core attention from $O(L^2)$ to $O(Lk)$; at a 128K context a query
attends to $64\times$ fewer keys~\cite{deepseek2025v32}.  DeepSeek-V4's
compressed sparse branch selects 512 or 1024 entries that each compress
four tokens, so at a one-million-token context a query covers about
$250$--$500\times$ fewer original positions; its heavily compressed branch
attends to all entries compressed 128-fold, which shrinks the quadratic
term by $128\times$ but keeps it quadratic,
$O(L^2/128)$~\cite{deepseek2026v4}.  

These figures count attended entries
or covered positions, not layer runtime, and omit indexing and
sliding-window branches.  What this paper relies on is the resulting layer
cost: in the profiled model and length range, sparse-layer time is
dominated by projections and MoE work, which are near-linear in length in
either layer type.  Only the attention term separates the two layer types,
so the inter-layer gap widens with sample length and is carried by the
long samples.  Wide groups are therefore sized by a minority of samples
for a minority of layers.

\subsection{Layerwise Topology Mismatch}
\label{sec:mismatch}

Wide SP16/32 groups are a product of the inter-sample gap: the baseline
solver allocates them only when a step's sample mix includes
samples whose full-attention work calls for wider parallelism, and the
same group then also hosts the shorter samples of that step.  The
inter-layer gap determines what such a group costs in the sparse layers.
Full attention benefits from wide SP16/32 because its quadratic attention and
activation state are distributed across more ranks.  For the longest
samples studied here, capping it at SP8 is no alternative: a group's
makespan cannot fall below its heaviest sample's cost on eight ranks, and
for these samples the quadratic full-attention term makes that cost far
exceed the group's average per-rank load, so full attention would run badly
imbalanced.  Sparse layers are different: for most evaluated workloads,
their near-linear MoE-dominated work already balances within node-local
SP8, while moving Ulysses to SP16/32 crosses the node boundary.

\begin{figure}[t]
  \centering
  \includegraphics[width=\linewidth]{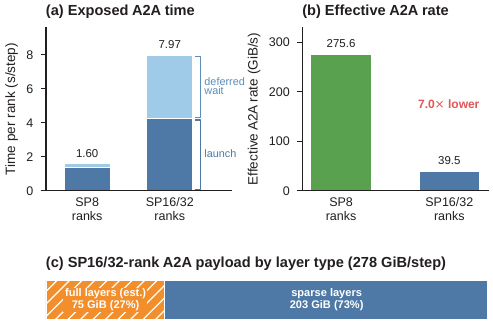}
  \caption{Ulysses communication on wide and node-local ranks for the
evaluated workloads. Unless noted otherwise, measurements use the
baseline solver with repacking disabled.
(a) Exposed all-to-all time per rank and step is 7.97\,s for SP16/32
ranks and 1.60\,s for node-local SP8 ranks.
(b) The corresponding effective rates are 39.5 and 275.6\,GiB/s.
Both measures include full-layer exchanges.
(c) Wide-rank payload by layer type; the full-layer share is estimated
from the conversion payload measured with repacking enabled.}
  \label{fig:comm-anti-scaling}
\end{figure}

We define exposed all-to-all time as the time a rank spends in all-to-all
calls, comprising launch time and deferred wait time; the latter includes
waiting for slower peers. A rank's effective rate is
its all-to-all payload divided by that time; because slower ranks make
others wait inside the collective, we take per step the rate of the rank
that waited least and average over steps.  Measured with the same
timer and estimator in the same run, the effective Ulysses rate is 275.6\,GiB/s on node-local SP8 ranks and 39.5\,GiB/s on
cross-node SP16/32 ranks, a $7.0\times$ gap; these are effective runtime
path rates rather than hardware link specifications.  Widening a sparse
block beyond the node therefore divides its computation over more ranks
but moves its exchange onto a much slower path.  Under
the baseline configuration (B0) of the evaluated workloads, wide SP16/32 ranks expose
7.97\,s of all-to-all per step versus 1.60\,s on node-local SP8 ranks
(Figure~\ref{fig:comm-anti-scaling}); all-to-all thus occupies 21.4\% of
the 37.22\,s wide-parent step on wide ranks, against 4.3\% on SP8 ranks.
These totals cover the Ulysses exchanges of all layers. A wide rank transfers 278\,GiB per step, of which approximately
75\,GiB is attributable to full layers, as estimated from the
conversion payload measured with repacking enabled
(Section~\ref{sec:comm-reduction}; Figure~\ref{fig:comm-anti-scaling}(c)).
The remaining approximately 203\,GiB is attributable to sparse layers.
This motivates retaining wide SP groups for full attention while moving
sparse-layer exchanges to node-local groups, thereby avoiding their
cross-node execution.

\subsection{Switching Layouts Moves Token Ownership}
\label{sec:ownership}

Node-local groups for the sparse layers are not a change of degree on the
same tokens.  An SP16 parent spreads its tokens over sixteen ranks in two
nodes, whereas an SP8 child holds whole samples inside one node, so moving
between the layouts changes which ranks own each sample's tokens.  A 1F3S
block crosses this boundary twice per full layer, into and out of full
attention, in the forward, recomputation, and backward passes.  Realized as
its own phase, each crossing would move the converted activations across
nodes in addition to the full layer's own Ulysses exchange.  On measured
payloads and rates, such a dedicated pass would cost about 0.9--1.9\,s per
step depending on whether it moves the hidden state or the attention
tensors, roughly 30--60\% of the exposed-all-to-all reduction that repacking
achieves (Section~\ref{sec:comm-reduction}).  Changing layouts per layer
type pays off fully only if it avoids that second transfer.

\subsection{Parent Composition Limits Child Balance}
\label{sec:composition}

Repacking balances work within a parent, so the balance it can reach is
limited by which samples share the parent, a decision made in planning.
Parent membership is chosen by the baseline solver's scalar weight, an
empirical score fitted to full-attention compute profiles. This approach is consistent with dynamic sequence-parallel schedulers
for homogeneous Transformer stacks, which assign each sample a profiled
cost under a shared parallel layout, as in FlexSP~\cite{wang2025flexsp}
and ByteScale~\cite{ge2025bytescale}.  When layers have similar cost
structures and use the same parallel groups, balancing aggregate
sample costs provides a basis for balancing group workloads.

A 1F3S hybrid stack
under repacking breaks the premise in two ways.  First, each sample now
carries a two-component compute cost, quadratic full-layer work and
near-linear sparse-layer work, and the baseline's weight, calibrated to
full attention, misjudges how unequal samples are: a long sample worth many
short ones under the quadratic term is worth only a few on the near-linear
sparse path.  Summing the two components into one per-sample total would
fix the totals but not the balance, because the two layer types now run on
different groups: full attention is balanced over the parent, whereas a
sparse layer finishes with its slowest child.  Membership that equalizes a
per-sample total can therefore sit far from balance on one layer class's
critical path.  Second, the empirical weight absorbs no communication cost;
the omission is benign while groups stay node-local, but the large samples
studied here force cross-node SP16/32 groups, where the un-priced
communication share becomes first-order.  Repacking removes that
cross-node term structurally (Section~\ref{sec:design}); the two compute
components remain for the planner.  The resulting mis-composition is visible
in both layouts.  A wide-parent step also contains ordinary SP8 parents,
which serve as a reference: with baseline membership and repacking
disabled, the forward rank max of the SP16/32 parent region exceeds that of
the ordinary SP8 ranks by 2.418\,s.  This imbalance motivates reassigning samples among parents while
preserving the parent topology, using a cost model that accounts for
both full- and sparse-layer computation. Section~\ref{sec:solver-gain}
indicates that the imbalance persists, albeit to a lesser extent, when
repacking is enabled.

\subsection{Why Pure Ulysses rather than USP}
\label{sec:whyup}

Pure Ulysses parallelism (pure UP) transposes sequence shards and attention
heads with all-to-all.  USP adds a Ring-Attention dimension so ring communication can
overlap attention computation~\cite{fang2024usp}.  In this workload, sparse
attention provides too little computation for effective overlap, and the
following MoE work lies outside that overlap window.  We therefore build on
pure UP as a design choice rather than a measured comparison with USP: it
avoids a parallel dimension whose overlap we expect to bring little benefit
in this workload, and
its mandatory all-to-all can also carry sample repacking
(Section~\ref{sec:design}).  Ring-based overlap of the remaining
full-layer exchange is orthogonal to repacking and could be combined with
it; we leave that combination to future work.

\section{\textsc{Layerwise Repacking} Design}
\label{sec:design}

\textsc{Layerwise Repacking} separates three decisions that the baseline couples: the
wide parent topology required by full attention, the node-local child layout
utilized by the sparse trunk, and the sample permutation between those layouts.
This section first formalizes the problem and its layer-specific layouts, then
describes a communication-fused transition and its correctness
requirements.

\subsection{Problem Formulation}
\label{sec:problem}

Consider one training step on a data-parallel world whose ranks form
eight-GPU nodes, and let sample $i$ have packed length $L_i$.  The baseline
solver (round~1A) partitions the world into contiguous parent buckets
\[
  b=(G_b,R_b,I_b), \qquad R_b=|G_b|\in\{8,16,32\},
\]
where $G_b$ is a rank interval and $I_b$ is its ordered sample list; every
rank in $G_b$ holds an identical assignment record.  With this parent
topology $(G_b,R_b)$ held fixed, we choose (i) the membership $I_b$ of every
parent and (ii) for each SP16/32 parent a set of children
\[
  \mathcal{P}_b=\{(G_{bj},r_{bj},I_{bj})\}_j,\qquad r_{bj}\in\{1,2,4,8\}.
\]
Here $G_{bj}$ is the rank interval assigned to child $j$,
$r_{bj}=|G_{bj}|$ is its sequence-parallel degree, and $I_{bj}$ is its
ordered sample list.  The child groups tile $G_b$, each inside one
eight-GPU node, and their non-empty sample lists partition the parent's
repaired sample list.  The sample assignment evolves through four stages:
\[
  I_b^0\;\rightarrow\;I_b^\star\;\rightarrow\;\widetilde I_b\;\rightarrow\;\{I_{bj}\}_j
\]
Here $I_b^0$ is the baseline assignment, $I_b^\star$ is the assignment
after optional parent reallocation (Section~\ref{sec:planner}), and
$\widetilde I_b$ is the assignment after repair
(Section~\ref{sec:runtime}); $\{I_{bj}\}_j$ is the resulting child
partition.  Repair may transfer small samples between parents to ensure
that non-empty children can be formed, but no such transfers were needed
in the evaluated workloads ($\widetilde I_b=I_b^\star$).  Child partitioning
alone is restricted to samples within a fixed parent.  Parent rank boundaries stay fixed throughout, and a parent
that cannot reach $R_b/8$ samples stays unsplit.  Full attention runs on the parents (layout
\layoutF{}) and the sparse layers on the children (layout \layoutS{}).  The
objective is the step's critical-path time, set by the slowest rank's
computation and exposed communication, subject to the requirement that
moving between the two layouts adds no redistribution pass; per-rank memory
is not modeled explicitly and is checked empirically
(Section~\ref{sec:layout-formulation}).  Section~\ref{sec:runtime} solves
(ii) for a given membership, and Section~\ref{sec:planner} improves (i)
with the topology held fixed.

\subsection{Layer-Specific Layout Formulation}
\label{sec:layout-formulation}

The full layout \layoutF{} keeps each parent's fixed rank interval $G_b$
with its repaired membership $\widetilde I_b$.  After the Q/K/V Ulysses
exchange, one rank holds the full parent token stream for its corresponding head slice,
approximately
\[
  X_{\layoutF}\in\mathbb{R}^{L_b\times(H/R_b)\times d}.
\]
Here $X_{\layoutF}$ denotes the rank-local attention tensor
in the full layout, with the parent token stream and one rank's share of
the attention heads.
The sparse layout \layoutS{} uses the rank's shard of one child stream,
\[
  X_{\layoutS}\in
  \mathbb{R}^{\lceil L_{bj}/r_{bj}\rceil\times H\times d}.
\]
Here $X_{\layoutS}$ denotes the rank-local tensor in the
sparse layout; the ceiling gives its padded token-shard length.
Here $L_b=\sum_{i\in\widetilde I_b}L_i$ and $L_{bj}=\sum_{i\in I_{bj}}L_i$
are the packed lengths of the parent and of child $j$, $H$ is the head
count, and $d$ is head width.  \layoutF{} exists only inside full
attention, between its two Ulysses exchanges; the projections and MoE work
of a full layer, like all sparse-layer work, run on \layoutS{} shards.
Padding is local to each child.  The correctness contract is that all \layoutS{} shards in
a parent form a disjoint partition of the parent's real packed tokens, and that
the inverse permutation reconstructs the original \layoutF{} packing order.
For token-proportional state, moving a sample from an SP$R_b$ parent to an
SP$r_{bj}$ child can increase that sample's per-rank residency by up to
$R_b/r_{bj}$ relative to \layoutF{}: SP32$\rightarrow$SP8 is a $4\times$
case, while the most concentrated SP32$\rightarrow$SP1 case is $32\times$.
This is a per-sample analytical bound; whole-device peaks are a different
metric, which the bound does not predict directly.  In the
four-configuration runs of Section~\ref{sec:evaluation}, worst-rank active
memory peaks at 71\% and reserved memory at 86--90\% of per-rank device
memory; enabling repacking shifts either peak by at most 0.5\% of capacity,
and no allocator retry occurs in any configuration (per-step worst over all
ranks).

\subsection{Communication-Fused Repacking}

For a repacked parent, the existing full-layer Q/K/V Ulysses site gathers the
node-local \layoutS{} shards into the wide \layoutF{} parent while performing
the standard sequence-to-head transpose.  The existing output site applies the
inverse sample permutation while returning to \layoutS{}.
Concretely, write $R=R_b$, assume $R$ divides $H$, and let the children give
parent rank $q$ the ordered set $T_q$ of $n_q$ real token positions in the
parent packing order.  Here $p,q\in\{0,\ldots,R-1\}$ index
ranks within the parent, $T_q\subseteq\{0,\ldots,L_b-1\}$, and
$n_q=|T_q|$.  At the Q/K/V site, rank $q$ sends to every parent
rank $p$ its tokens restricted to head slice $p$,
\[
  X\bigl[T_q,\ pH/R:(p+1)H/R,\ :\bigr],
\]
so its send splits are $n_q$ tokens to every destination and its receive
splits are $(n_0,\dots,n_{R-1})$.  The received stream is ordered by source
rank, $T=(T_0,\dots,T_{R-1})$, and the permutation $\pi=\operatorname{argsort}(T)$
indexes this rank-major stream in ascending parent token
order, yielding exactly the head-sharded input that full
attention consumes without repacking.  The output site inverts the map:
rank $p$ reorders its attention output by $T$, sends $n_q$ tokens to each
rank $q$, and each $q$ reassembles all heads for $T_q$.  Q, K, and V each
pass through the Q/K/V site, separately for each modality stream.  With
uniform shards, $T_q$ is the $q$-th contiguous chunk, $\pi$ is the
identity, and both sites reduce to standard Ulysses.

Generalizing the split sizes therefore carries repacking inside
communication that the full layer already requires.  Every tensor still
traverses each mandatory exchange once; only split sizes and the local
permutation change, so no payload is transferred twice.  The absence of an additional pass means that repacking requires no
separate inter-rank redistribution; the local permutation still uses
buffers, and the full layer retains its mandatory Ulysses exchanges. The performance gain comes from keeping the
three sparse layers node-local.

\subsection{What the Fused Exchange Must Preserve}
\label{sec:fusion-requirements}

Reusing the mandatory exchange is a correctness burden rather than a free
generalization; variable split sizes deliver repacking only if four
properties hold simultaneously.  First, the forward and reverse layout transitions must be exact
inverses. The \layoutS{}$\rightarrow$\layoutF{} gather must map each
child-shard token to its unique position in the parent packing order,
and the \layoutF{}$\rightarrow$\layoutS{} return must restore the
original child-shard assignment. This establishes the disjoint-partition
and inverse-permutation requirements of
Section~\ref{sec:layout-formulation}.  Second, every rank must
derive the permutation independently and identically: repacking metadata
cannot itself become a new collective, so all ranks build the same
deterministic rank-major index tables from the shared plan.  Third, padding
and packed-length bookkeeping must stay local to each child so
that existing attention and MoE kernels see ordinary packed inputs.  Fourth,
the backward and recomputation graphs must traverse the same uneven
exchanges in reverse with consistent split metadata, and the loss path must
reconstruct global packed order without adding a phase.
Section~\ref{sec:runtime} realizes these requirements.

\section{Runtime Realization}
\label{sec:runtime}

This section realizes the design of Section~\ref{sec:design} for a given
parent membership, produced by either solver.  The runtime first repairs
sample counts, plans the children, and constructs deterministic token
indices.  At each full layer, the existing Ulysses Q/K/V and output
exchanges temporarily realize \layoutF{} and return to \layoutS{}.

\subsection{Child Planning and Repair}

For sample feature vector $\phi_i$, a calibrated sparse-work predictor gives
the sparse-layer work score
\begin{equation}
  s_i=\mathcal C_{\mathrm{sparse}}(\phi_i;\theta_S),
\label{eq:sparse-cost}
\end{equation}
a deterministic score summarizing sparse-attention and near-linear
MoE-dominated per-token work.  Here $\phi_i$ contains the
profiled features of sample $i$, $\mathcal C_{\mathrm{sparse}}$ is the
sparse-layer work predictor, and $\theta_S$ denotes its calibrated
parameters.  The child planner approximately minimizes,
for each repaired parent,
\begin{equation}
  \begin{aligned}
  \min_{\mathcal{P}_b}
  \max_j & \frac{\sum_{i\in I_{bj}}s_i}{r_{bj}}, \\
  \text{s.t.}\quad
  \sum_j r_{bj}=R_b,&\quad r_{bj}\in\{1,2,4,8\},\quad I_{bj}\ne\emptyset,
  \end{aligned}
\end{equation}
where each $G_{bj}$ is an interval of $r_{bj}$ ranks aligned to a multiple
of $r_{bj}$ within the parent, so every child lies inside one eight-GPU
node.
It reuses the SP-size allocator of round~1A, run in pure-Ulysses
mode inside the parent with the four child sizes as candidate degrees.

Before child planning, the runtime repairs parent memberships to ensure
that every rank can be assigned to a non-empty, node-local child group
of at most eight ranks. This requires
\[
  |I_b|\ge m_b,\qquad
  m_b=
  \begin{cases}
    R_b/8,& R_b\ge16,\\
    1,& R_b<16,
  \end{cases}
\]
Here $m_b$ is the minimum number of samples required to
form non-empty node-local children for parent $b$; the constraint applies
to its membership after repair.  Thus, an SP16 parent needs at least two samples and an SP32 parent at least
four.  For a deficient parent, it selects the donor with the largest predicted
per-rank sparse cost among donors with surplus, then moves that donor's
smallest-cost sample.  Donors retain their own minimum.  Ties follow rank and
sample order, so identical inputs produce identical plans.

Repair changes sample membership and total length while preserving parent rank
intervals.  The child planner then operates on the repaired parent pools.
Repair runs only when repacking is enabled, so the repacking-only effect in
Section~\ref{sec:evaluation} includes it; in the evaluated workloads every wide
parent already met its minimum and repair moved no sample.

After repair, child planning follows one of three execution paths:
\begin{itemize}
  \item \textbf{Primary.} The child planner returns non-empty children and
  the result is materialized.
  \item \textbf{Uniform SP8.} An exception or empty-rank result triggers greedy
  assignment into $R_b/8$ uniform SP8 bins.  This preserves node locality.
  \item \textbf{Unsplit.} SP8 parents are unchanged by design.  An SP16/32
  parent also remains unsplit if repair cannot reach $R_b/8$ samples.
\end{itemize}

\subsection{Sparse-Trunk Execution}

The route carries the parent rank start, parent SP, repaired sample order, and
child descriptions.  At model entry, every rank builds the same
rank-major token-index table for each modality.  The table serves three roles:
it selects the rank's input shard, permutes an \layoutS{} receive stream back
into \layoutF{} order, and gathers an \layoutF{} output into destination-major
\layoutS{} order.
Model inputs are globally replicated at this point, so selecting the initial
\layoutS{} shard is a local indexed selection; it is not a network
collective.  Hidden states, residuals, masks, time conditioning, positional
metadata, and attention-residual history remain in \layoutS{} for the complete
trunk.  Sparse layers receive child-specific packed lengths and run their
existing attention stack under the subgroup of size $r_{bj}$.

Every repacked child has SP$\le8$ and remains inside one eight-GPU node, so
no sparse layer runs a parent-wide cross-node attention Ulysses exchange.
Sparse layers use
the normal Ulysses primitive within the child group when $r_{bj}>1$; an SP1
child needs no attention Ulysses exchange, although other collectives of
the training step remain.  The near-linear MoE-dominated work is thus approximately balanced
over local ranks while its exchange stays on the intra-node fabric.

Let $N_F$ and $N_S$ be the numbers of full and sparse layers.  Counting the two
pre/post-attention Ulysses sites in the forward graph gives
\[
  N_{\mathrm{xnode}}^{\mathrm{base}}=2(N_F+N_S),
  \qquad
  N_{\mathrm{xnode}}^{\mathrm{repack}}=2N_F.
\]
Here $N_{\mathrm{xnode}}^{\mathrm{base}}$ and
$N_{\mathrm{xnode}}^{\mathrm{repack}}$ count cross-node Ulysses call sites
per repacked parent's forward pass under the baseline and repacked layouts,
respectively.
A 1F3S stack has $N_S=3N_F$, so for a repacked parent the cross-node
Ulysses call sites fall from $8N_F$ to $2N_F$, a 75\% structural reduction.
This count describes the forward graph; backward and recomputation preserve the
same layer-type ratio.

\subsection{Communication-Fused Execution}

For a repacked parent, a full layer performs:
\begin{enumerate}
  \item \textbf{\layoutS{}$\rightarrow$\layoutF{} on Q/K/V.}  Each rank splits
  heads by parent destination, executes an uneven all-to-all over $G_b$, and
  permutes the rank-major receive stream into parent packing order.
  \item \textbf{Full attention in \layoutF{}.}  Each rank owns all parent
  tokens for $H/R_b$ heads.
  \item \textbf{\layoutF{}$\rightarrow$\layoutS{} on output.}  The parent
  stream is gathered in destination order, exchanged with uneven splits, and
  reshaped to all heads on the local \layoutS{} token shard.
\end{enumerate}

Steps 1 and 3 are the original Ulysses call sites with the split sizes and
permutation of Section~\ref{sec:design}.

After the trunk, an uneven all-gather reconstructs each active modality in
global packed order for the loss path.  It replaces the baseline uniform
all-gather rather than adding a new loss-path phase.  Parents with insufficient
samples retain their original layout.

\section{Improved Parent Solver}
\label{sec:planner}

The improved solver is an auxiliary planner: it does not change how
repacking executes, only the parent membership repacking starts from.

\subsection{Why Membership Matters}

Within-parent runtime repacking changes how a wide parent executes sparse
layers, but not which samples the parent contains.  Repair moves only enough samples across
parents to make children feasible, and the child planner then partitions
each parent independently, so it cannot balance a heavy parent with light
samples held by another parent.  This imposes a hard lower bound: with the
sparse-work score $s_i$ of Equation~\eqref{eq:sparse-cost}, the best
achievable sparse makespan of parent $b$ satisfies
\begin{equation}
  M_b^\star \ge
  \max\left(
    \frac{\sum_{i\in \widetilde I_b}s_i}{R_b},
    \max_{i\in \widetilde I_b}\frac{s_i}{8}
  \right).
\label{eq:child-lower-bound}
\end{equation}
Here $M_b^\star$ denotes the minimum predicted sparse-layer
makespan over feasible child partitions of parent $b$.
Under this per-sample cost model, the first term is the ideal average over
the parent; the second follows from
the child-SP cap $r_{bj}\le8$: even the heaviest sample can use at most eight
ranks.

The improved solver (round~1B) targets the first term of this bound.  The
second depends on membership only through which parent holds each heavy
sample: moving a heavy sample lowers its old parent's term, but the
step's heaviest sample sets the term in whichever parent receives it, and
no membership change removes that bound (Section~\ref{sec:limitations}).
Before
child planning, it keeps each parent topology $(G_b,R_b)$ fixed while
reassigning sample membership across parents.  We write $I_b^0$ and
$I_b^\star$ for the membership after rounds~1A and~1B.  Repacking remains the
communication and layout mechanism; round~1B improves the children it can
produce.

\begin{table}[t]
  \caption{Planner stages and optimization domains.}
  \label{tab:planner-stages}
  \small
  \begin{tabularx}{\columnwidth}{@{}p{0.22\columnwidth}p{0.28\columnwidth}X@{}}
    \toprule
    \textbf{Stage} & \textbf{Fixed} & \textbf{Optimized output} \\
    \midrule
    Round-1A baseline &
      DP world, allowed SP &
      Seed topology $(G_b,R_b)$ and membership $I_b^0$ \\
    Round-1B improved &
      All $(G_b,R_b)$ from round-1A &
      Global parent membership $I_b^\star$ \\
    Repair &
      All $(G_b,R_b)$ &
      Feasible membership $\widetilde I_b$ ($|\widetilde I_b|\ge m_b$) \\
    Round-2 child planner &
      $(G_b,R_b)$ and $\widetilde I_b$ &
      Per-parent child groups $(G_{bj},r_{bj},I_{bj})$ \\
    \bottomrule
  \end{tabularx}
\end{table}

Table~\ref{tab:planner-stages} summarizes the planning stages, their
fixed constraints, and their outputs. Execution follows Table~\ref{tab:planner-stages}: round~1A fixes the parent
topology, optional round~1B changes only parent membership, repair
restores feasibility, and round~2 forms the node-local children. Round~1B optimizes sample membership while preserving the parent
topology $(G_b,R_b)$, thereby restricting the search space to sample
reassignment. The round-1A assignment is retained as a fixed fallback
if no route yields a valid candidate; it is not scored against the
generated candidates. The baseline
and improved evaluation configurations use the same repair and round-2
implementation; they differ only in whether round~1B runs.

\subsection{Parent Reallocation}

\begin{table*}[t]
  \centering
  \caption{Performance on all 22 SP16/32 workloads.  Values average the 22
  workloads; B0$\rightarrow$N1 is the complete-system comparison.
  All reported changes are calculated from unrounded means.}
  \label{tab:performance}
  {\small
  \begin{tabular}{@{}llrrrr@{}}
    \toprule
    \textbf{Cell} & \textbf{Configuration} & \textbf{Mean E2E (s)} &
      \textbf{Mean MFU (\%)} & \textbf{Fwd max (s)} & \textbf{Bwd max (s)} \\
    \midrule
    B0 & baseline solver / OFF (original) &
      37.221 & 56.47 & 8.864 & 23.316 \\
    B1 & baseline solver / ON  & 35.007 & 60.22 & 7.437 & 22.302 \\
    N0 & improved solver / OFF & 35.887 & 58.52 & 8.241 & 22.766 \\
    \textbf{N1} & \textbf{improved solver / ON (complete)} &
      \textbf{34.390} & \textbf{61.30} & \textbf{7.218} & \textbf{21.900} \\
    \bottomrule
  \end{tabular}}
\end{table*}

\begin{figure*}[t]
  \centering
  \includegraphics[width=\textwidth]{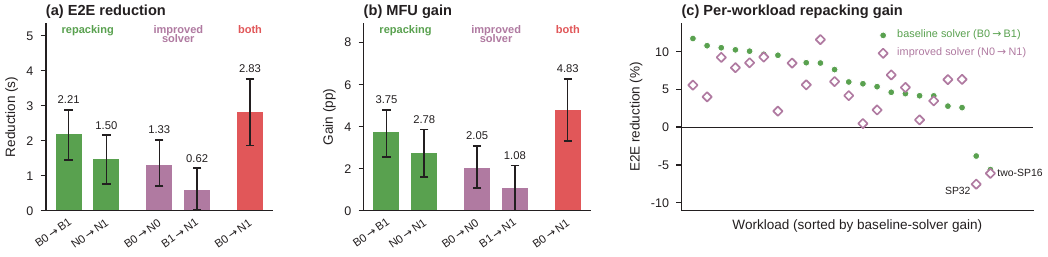}
  \caption{Results on the 22 SP16/32 workloads.  (a, b) Paired E2E reduction
  and MFU gain for the two repacking contrasts (B0$\rightarrow$B1,
  N0$\rightarrow$N1), the two solver contrasts (B0$\rightarrow$N0,
  B1$\rightarrow$N1), and the complete system (B0$\rightarrow$N1); bars are
  means over the 22 workloads and whiskers are workload-level 95\% bootstrap
  confidence intervals.  (c) E2E reduction from repacking on each workload,
  relative to the same solver without repacking.}
  \Description{Left two panels: bar charts with confidence intervals; E2E
  falls by 2.21 and 1.50 s with repacking, 1.33 and 0.62 s with the improved
  solver, and 2.83 s with both; MFU rises by 3.75, 2.78, 2.05, 1.08, and 4.83
  percentage points. Right panel: per-workload E2E reduction from repacking,
  positive on 20 workloads and negative on the two-SP16 and SP32 workloads
  under both solvers.}
  \label{fig:overall-performance}
\end{figure*}

Round~1B evaluates parent membership with a calibrated modality-aware predictor
rather than only the baseline scalar weight; the predictor is fitted to
profiled data per modality and per SP degree.  For sample $i$ and parent degree
$R$, the predictor returns a reference work score $a_i$ and
an SP-dependent marginal $m_i(R)$:
\begin{equation}
  a_i=\mathcal C_{\mathrm{ref}}(\phi_i;\theta_P),
  \qquad
  m_i(R)=\mathcal C_{\mathrm{marg}}(\phi_i,R;\theta_P).
\label{eq:sample-marginal}
\end{equation}
Here $\theta_P$ denotes the calibrated parameters of the
parent cost model; $\mathcal C_{\mathrm{ref}}$ and
$\mathcal C_{\mathrm{marg}}$ predict reference work and SP-dependent
incremental load, respectively.
Each greedy bucket starts with
\begin{equation}
  \ell_0(R)=\mathcal C_{\mathrm{base}}(R;\theta_P),
\label{eq:initial-parent-load}
\end{equation}
then accumulates $m_i(R)$.  Here $\ell_0(R)$ is the initial
load predicted by $\mathcal C_{\mathrm{base}}$ for an empty SP-$R$ parent.
The predictor combines modality, attention, dense,
and MoE forward/backward effects into deterministic scores used for allocation
ordering and comparison.

After allocation, the evaluator rebuilds each aggregate bucket and scores it
with an aggregate critical-path predictor rather than merely trusting the
marginal sum:
\begin{equation}
  \widehat T(F)=
  \mathcal C_{\mathrm{e2e}}
  \!\left(\{(I_b,R_b)\}_{b\in\mathcal B};\theta_P\right).
\label{eq:parent-eval}
\end{equation}
Here $\mathcal B$ is the fixed set of parents,
$F=\{(I_b,R_b)\}_{b\in\mathcal B}$ is a candidate assignment, and
$\mathcal C_{\mathrm{e2e}}$ predicts its step-level critical-path time
$\widehat T(F)$.
The evaluator is a parent-level surrogate for the slowest-rank symptom,
not average work; it does not model repair or child partitioning.  It scores the membership $I_b^\star$ before repair and
child planning.  Repair moved no sample in the evaluated workloads, so the
scored and executed memberships coincide there; the executed child plan
is still not scored (Section~\ref{sec:limitations}).

\subsection{Candidate Portfolio and Selection}

The solver generates a small portfolio of candidate memberships rather than
optimizing one weighted compute--memory objective.  Two resource-priority
routes, \textsc{Compute-LPT} and \textsc{Memory-LPT}, are two priority
views of one greedy allocator: they order samples by SP-dependent marginal
work or by packed length, pre-place required heavy samples in
distinct wide parents, and place each remaining sample where the projected
global makespan is smallest, under the baseline per-parent length cap.  An
optional \textsc{Anchor-Fill} route instead shapes each
wide parent explicitly for the downstream child planner, combining
anchor/filler decomposition~\cite{xiang2026zellige} with classical
large/small-job organization~\cite{hochbaum1987dual}.  Every valid
candidate is rebuilt from its sample lists and scored by
Equation~\eqref{eq:parent-eval}; selection takes the tolerance-tied minimum
of $\widehat T(F_r)$ with a fixed route order as tie break.
Here $r$ indexes the candidate-generation routes, and $F_r$
is the assignment produced by route $r$.  The evaluator
thus chooses among routes per workload, and the evaluation reports the
complete portfolio as one improved-solver configuration.
Appendix~\ref{app:solver-routes} specifies the queue keys, placement rule,
budget, selection equation, and pseudocode.

\paragraph{Solver cost.}
Each route sorts the $n$ samples once and places each sample with a constant
number of balanced-tree and heap queries per SP class, so one route costs
$O(n\log n+Cn\log P)$ for $P$ parents and $C\le3$ SP classes, and the
evaluator is linear in $n+P$.  The measured solve time is at most 4\,ms per
step.

\section{Experiments}
\label{sec:evaluation}

We evaluate the end-to-end performance of repacking alone and the
complete system, then analyze communication-time savings and the
incremental gains from the improved parent solver.

\subsection{Experimental Setup}

All runs use 64 GPUs on eight nodes (eight GPUs per node) with bf16
training; Section~\ref{sec:mismatch} characterizes the interconnect by its
measured effective rates.  We collected the first 200 training steps of
our 1F3S hybrid-attention model; 22 of them contain an SP16 or SP32 parent and therefore trigger
repacking.  We call each of these steps a workload and run all 22 under four
configurations, together with 20 SP8-only steps from the same window as
controls.  The configurations are labeled by parent solver and repacking state: B0 (baseline
solver, repacking OFF; the original system), B1 (baseline solver, ON), N0
(improved solver, OFF), and N1 (improved solver, ON; the complete system).
The baseline solver follows the design of Megatron-Core's Dynamic Context
Parallelism~\cite{li2026dynamiccp}: it chooses each packed micro-batch's
parallel degree from pre-created power-of-two groups that every layer then
shares; its groups run pure Ulysses here.  The child
planner uses SP$\{1,2,4,8\}$, while the improved solver preserves the
parent SP topology.  The set contains 20 one-SP16 workloads, one two-SP16 workload, and
one one-SP32 workload, so the evidence chiefly covers one-SP16 parents; the
two-SP16 and SP32 topologies are each represented by a single workload.
Means are taken over the 22 workloads, and 95\% confidence intervals use a workload-level bootstrap.  MFU follows the standard definition:
theoretical model FLOPs of the non-padding tokens (forward plus backward,
excluding recomputation) per device-second, divided by the device peak
FLOPs.  Because the four configurations run as separate jobs, the control
steps, on which neither lever changes the plan, provide a diagnostic
reference for cross-run machine-efficiency differences (after a configuration-blind
exclusion of one isolated stall): on identical SP8 plans, the
configuration means differ by at most 0.346\,s, about one eighth of the
2.83\,s headline gain.

We assess numerical consistency by comparing training
losses across the four configurations, using the same initial state and
training steps.  The first step, which precedes any wide parent, has bitwise
identical loss across configurations, confirming a common starting state.
Measured against B0, the mean absolute loss difference
$|\Delta\mathrm{loss}|$ over the wide-parent steps is $5.7\times10^{-4}$ (B1), $5.2\times10^{-4}$ (N0), and
$4.7\times10^{-4}$ (N1).  On the control steps, where no wide parent exists
and neither lever changes execution, it is already
$1.2$--$1.9\times10^{-4}$: nondeterministic reduction order makes
trajectories drift apart through optimizer updates, so later steps are not
expected to match bitwise.  On the first wide-parent step, where repacking
first changes the layout, $|\Delta\mathrm{loss}|$ is
$0.8$--$1.9\times10^{-4}$, within the range seen on the control steps
(up to $6\times10^{-4}$).  Loss agreement is a necessary check rather than
a proof; correctness rests on the bijection and reverse-path requirements
of Section~\ref{sec:fusion-requirements}, which the implementation is
designed to satisfy, rather than on a separate output- or
gradient-equivalence test.

\subsection{Overall End-to-End Performance}

Table~\ref{tab:performance} and Figure~\ref{fig:overall-performance} show
the results.  Repacking alone lowers E2E by 2.214\,s (5.95\%) with the
baseline solver and 1.497\,s (4.17\%) with the improved solver; the
corresponding MFU gains are 3.75 and 2.78\pp{}.
Figure~\ref{fig:overall-performance}(c) shows that these gains are not driven
by the mean alone: median per-workload E2E gains are 2.46 and 1.85\,s, and the
workload-level 95\% confidence intervals exclude zero for both E2E
(1.44--2.88\,s and 0.76--2.15\,s) and MFU
(2.55--4.79\pp{} and 1.61--3.85\pp).  Under both solvers, repacking lowers
mean E2E on all 20 one-SP16 workloads (by 2.6--11.7\% under the baseline
solver) and raises it on the two-SP16 and
SP32 workloads, where the SP8 child cap leaves one long sample dominating
its parent (Section~\ref{sec:limitations}).

With the improved solver, the complete system (N1 versus B0) lowers mean
E2E from 37.221 to 34.390\,s, a 2.831\,s (7.61\%) reduction with a
workload-level 95\% confidence interval of 1.86--3.77\,s, and raises MFU
from 56.47\% to 61.30\%, a 4.83\pp{} (3.31--6.27\pp{}) increase.  Forward
and backward rank-max time fall by 1.646 and 1.416\,s, respectively; the
two maxima need not come from the same rank, so they do not add up to the
E2E change.

Whole-window arithmetic makes the scope of that headline explicit.  Wide
parents arise only when a step's sample mix leads the baseline solver to
allocate SP16/32 groups, which happens on 22 of the 200 collected steps; the
other 178 steps run entirely on SP8 groups, which the two levers leave
untouched by construction: repacking splits only SP16/32 parents and the
improved solver is disabled on pure-SP8 topologies, so all four
configurations execute the same plan on these steps.  Estimating each of
them by the 28.27\,s B0 mean of the SP8-only control steps, wide-parent steps
(37.221\,s) are the slowest steps in the window and account for about 14\%
of its estimated wall-clock ($22\times37.221$\,s out of $\approx$5851\,s).
The 7.61\% headline is a per-triggered-step figure; the window-level
saving is approximately the wall-clock share of triggered steps times the
per-step gain, so it varies with how often a workload triggers wide
parents.  Under a given scheduler and constraints, the share can increase with the spread of sample lengths, since the
further a step's longest samples outgrow the rest, the more often the
baseline solver may need to allocate wide groups.

\subsection{Communication Reduction}
\label{sec:comm-reduction}

\begin{figure}[t]
  \centering
  \includegraphics[width=\linewidth]{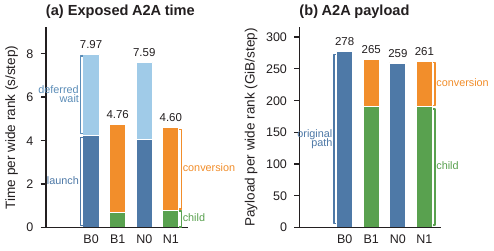}
  \caption{Communication on SP16/32 ranks before and after repacking.
  ``Original'' is the parent-wide Ulysses path without repacking; ``child''
  and ``conversion'' are the node-local sparse-layer exchanges and the
  fused full-layer exchanges.  Exposed A2A time falls
  from 7.97/7.59\,s to 4.76/4.60\,s under the baseline/improved solvers.
  Total payload changes by less than 5\%, and more than 70\% of the repacked
  payload uses the node-local child path.}
  \Description{Stacked bars of all-to-all time and payload on wide ranks
  before and after repacking under both solvers.}
  \label{fig:repack-comm-overhead}
\end{figure}

Figure~\ref{fig:repack-comm-overhead} shows the central mechanism at work:
repacking replaces sparse-layer parent-wide A2A with node-local child A2A
and uses the mandatory full-layer exchange for
\layoutS{}$\leftrightarrow$\layoutF{} conversion, so exposed A2A on wide
ranks falls by 3.20/2.99\,s while total payload stays nearly unchanged; most
traffic is relocated without a second payload traversal.

Time shares and rates show where the remaining communication sits.  On wide
ranks, all-to-all falls from 21.4\% to 13.6\% of the step under the
baseline solver and from 21.1\% to 13.4\% under the improved solver.  After
repacking, the node-local child exchanges carry more than 70\% of the
wide-rank payload yet take only 14--17\% of the remaining all-to-all time,
because they stay on the intra-node fabric (Section~\ref{sec:mismatch}).  The
other 83--86\% (4.08/3.82\,s) is the mandatory full-layer
\layoutS{}$\leftrightarrow$\layoutF{} exchange, which any wide-SP full
attention must send across nodes.  For repacked parents, the residual
communication is therefore the traffic full attention itself requires; the
avoidable cross-node sparse-layer traffic is gone.

A dedicated redistribution pass is the natural alternative realization, so
Table~\ref{tab:dedicated-pass} estimates the pass that fusion avoids.  The
conversion streams carry 74.6/70.5\,GiB per wide rank per step under
the baseline/improved solvers.  The cost of a separate pass depends on what
it moves.  A pass that converts the same attention tensors (Q, K, and V on
entry, the output on exit) would send that payload across nodes a second
time, about 1.89/1.78\,s per step at the observed 39.5\,GiB/s.  A pass that
instead moves the hidden state before the Q/K/V projection and after the
output projection sends two tensors of the same width as each attention
tensor, at the same precision, rather than four, halving the bytes to
about 0.94/0.89\,s.  Either way the pass would return roughly 30--60\% of
the 3.20/2.99\,s exposed-A2A reduction that repacking achieves.  Across the
other observed rates---least-blocked conversion rates of 42.9/42.5\,GiB/s
(per step, the payload/time ratio of the rank that waited least inside its
collectives, averaged over steps) and as-deployed wide-path rates of
34.9/34.2\,GiB/s---the estimates span 1.7--2.1\,s and 0.8--1.1\,s for the
two designs.  Avoiding a separate redistribution pass is therefore a
saving, estimated from measured payloads and rates at about 0.4--1.2 times
the repacking E2E gain itself (2.214/1.497\,s) at 39.5\,GiB/s, and not only a structural
property of the design.

\subsection{Parent-Solver Gains}
\label{sec:solver-gain}

Parent-composition imbalance predates repacking
(Section~\ref{sec:composition}).  With baseline membership,
the parent-region forward rank max exceeds ordinary SP8 by 2.418\,s with
repacking OFF and 0.696\,s with repacking ON; improved membership lowers the
same gaps to 1.692\,s and 0.452\,s.  Each value averages the 23 wide
parents of the 22 workloads; the two-SP16 workload contributes two.  Round~1B thus repairs a planning-level residual
that exists in both layouts, rather than a condition repacking creates.

With repacking OFF, B0$\rightarrow$N0 lowers E2E by 1.334\,s
(Figure~\ref{fig:overall-performance}(a)), raises MFU by
2.05\pp{}, and lowers forward/backward rank max by 0.624/0.550\,s.  With
repacking ON, B1$\rightarrow$N1 lowers E2E by 0.617\,s, raises MFU by
1.08\pp{}, and lowers the two rank maxima by 0.219/0.402\,s.  The E2E
confidence intervals are 0.70--2.02\,s and 0.02--1.21\,s, respectively; the
near-zero lower end with repacking ON makes this residual gain less certain
than the repacking effect.  The gain is also not uniform: in each repacking
state, the improved solver does not lower mean E2E on 7 of the 22 workloads, so its
benefit is an average over workloads rather than a per-step guarantee.

The two levers overlap rather than add: the solver's
E2E gain is larger with repacking OFF (1.334\,s) than ON (0.617\,s), and
the combined B0$\rightarrow$N1 improvement of 2.831\,s falls 0.717\,s short
of the sum of the individual effects.  Two mechanisms explain the overlap:
repacking structurally removes the un-priced cross-node communication share
(Section~\ref{sec:composition}), and its child planner, which places samples
of unequal cost within each parent, absorbs part of the mis-membership, so
less residual imbalance remains for the improved solver.

\begin{table}[t]
  \centering
  \caption{Estimated redistribution overhead avoided by
communication fusion, per wide rank and step (forward, recomputation,
and backward combined). Conversion payload and A2A time savings are
measured. Estimates assume four attention-tensor transfers
(Q/K/V in; output out) or two hidden-state transfers (in/out) per full
layer and phase at the measured cross-node effective rate of
39.5\,GiB/s, excluding additional launch and synchronization overheads.
The fraction is estimated overhead divided by measured A2A time savings.}
  \label{tab:dedicated-pass}
  {\small
  \begin{tabular}{@{}lrr@{}}
    \toprule
 \textbf{Metric}   & \textbf{Baseline} & \textbf{Improved} \\
    \midrule
    \makecell[l]{Conversion payload (GiB)} & 74.6 & 70.5 \\
    \addlinespace
    \makecell[l]{Pass moving attention tensors (s)} & $\approx$1.89 & $\approx$1.78 \\
    \addlinespace
    \makecell[l]{Pass moving hidden state (s)} & $\approx$0.94 & $\approx$0.89 \\
    \addlinespace
    \makecell[l]{A2A reduction from repacking (s)} & 3.20 & 2.99 \\
    \addlinespace
    \makecell[l]{Fraction of reduction lost to the pass} & 30--59\% & 30--60\% \\
    \bottomrule
  \end{tabular}}
\end{table}

\begin{figure}[t]
  \centering
  \includegraphics[width=\linewidth]{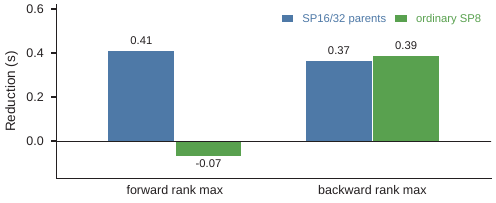}
  \caption{Where the improved solver's gain lands: rank-max reduction in
  SP16/32 parent and ordinary-SP8 regions, equally averaging the repacking-OFF
  and -ON deltas over the same 23 wide parents.}
  \Description{Grouped bars: forward rank max falls by 0.41 s in SP16/32
  parents and rises by 0.07 s in ordinary SP8 regions; backward rank max falls
  by 0.37 and 0.39 s.}
  \label{fig:solver-gain}
\end{figure}

The forward-timing reduction is concentrated where the solver changes
membership.  Across both repacking states, parent-region forward rank max
improves by 0.414\,s, against $-0.071$\,s in the ordinary-SP8 region
(Figure~\ref{fig:solver-gain}).  Backward reduction is similar in the two
regions (0.367 and 0.389\,s), indicating that the solver's localized effect appears primarily in
the forward parent composition.

\section{Limitations and Future Directions}
\label{sec:limitations}

This section discusses the performance and applicability
limits of \system{}, including the SP8 child cap, evaluation scope,
and memory and calibration constraints. We then outline potential
extensions for benefit-aware repacking, adaptive child-group sizing,
and execution-plan evaluation.

\paragraph{The SP8 child cap.}
The current version caps every child at SP8 so that no sparse-attention
Ulysses exchange crosses a node.  The cap meets the same single-sample bound that rules out SP8 for
full attention (Section~\ref{sec:mismatch}), only far less often because
sparse-layer work is near-linear.  When a few extremely long samples dominate
their parent, it bounds the parent's sparse makespan at the heaviest sample's SP8 cost (the
second term of Equation~\eqref{eq:child-lower-bound}), and the resulting
load imbalance can consume the communication saving.  This happens on
exactly the two-SP16 and SP32 workloads, the only ones on which repacking
raises mean E2E (by up to 3.3\,s): measured by length, their heaviest sample
spread over eight ranks carries 1.47--1.70$\times$ its parent's average
per-rank load, more than in any improving workload.  The current version
therefore does not improve the SP32 step.

\paragraph{Scope of the evaluation.}
All results come from one 1F3S model on one cluster and cover the 22
wide-parent steps among the first 200 collected training steps.  The length
distribution, and hence how often wide parents arise, may change later in
training.

\paragraph{Applicability.}
The eight-GPU node here is one instance of a more general boundary: the
high-bandwidth domain within which all-to-all runs fast.  A larger domain
postpones the mismatch, but it reappears whenever full attention needs a
group wider than the domain.  Nor is the mechanism tied to sparse
attention: other layers whose work is near-linear in length, such as
linear attention, are candidates for node-local children while full
attention keeps its wide parent, provided their state, communication, and
operator structure permit the same split; we have not validated this.

\paragraph{Memory and calibration.}
Repacking can raise a sample's per-rank activation residency by up to
$R_b/r_{bj}$ (Section~\ref{sec:layout-formulation}); the evaluated steps
show little change in whole-device peaks, but steps with larger samples may
behave differently.  The improved solver relies on a calibrated cost
predictor, which must be recalibrated when the model or cluster changes.

\paragraph{Future directions.}
Three extensions address these limits directly.  First, repacking could be
gated by predicted gain rather than by feasibility alone.  The fallback
policy handles plans that cannot be built, but the two regressing workloads
were feasible and slower.  Keeping the unsplit parent as an explicit
candidate, and repacking only when the predicted gain exceeds a margin,
would target them; the second term of
Equation~\eqref{eq:child-lower-bound} offers a cheap trigger: measured by
length, the heaviest sample's eight-rank load is 1.47--1.70 times its
parent's average per-rank load in the regressing workloads, against at
most 1.34 in improving ones, although a threshold drawn from these 22
workloads needs validation elsewhere.  Second, the child cap could become a
choice: for an extremely long sample, the planner would compare an SP8
child, an SP16 child that crosses one node boundary, and keeping the
parent, while the remaining samples stay node-local; the objective is the
step's critical path rather than node locality of every sparse exchange.
Third, the solver could score the executed plan.  Its evaluator currently
scores membership before repair and child planning; scoring each candidate
after repair, child planning, and feasibility checks---including the
slowest child's time, full- and sparse-layer communication, permutation and
padding overhead, and peak memory---would close the gap between the scored
and the executed plan and make memory an explicit constraint.

\section{Related Work}

\subsection{Sequence Parallelism}
Sequence parallelism partitions token activations across devices to extend
context length and compose with other parallel
dimensions~\cite{li2023sequence}.  Ulysses realizes the sequence/head transpose
with all-to-all~\cite{jacobs2023deepspeed}; Ring Attention streams blockwise
key/value states around a device ring~\cite{liu2024ringattention}; USP combines
the two dimensions~\cite{fang2024usp}; LoongTrain's 2D-Attention likewise
combines head- and context-parallel groups and studies their device
placement~\cite{gu2024loongtrain}; and DSP switches the sharded sequence
dimension across computation stages~\cite{zhao2025dsp}.  DSP is the closest
transition mechanism: it also moves tokens with all-to-all between stages,
but keeps each sample's participating rank set and the group width fixed,
so its splits are uniform.  \textsc{Layerwise Repacking} instead changes a sample's
participating ranks and group width per layer type, moving it between a
wide parent and a node-local child, which requires variable split sizes
and an explicit sample permutation inside the same exchange.  DCP
partitions attention data and computation into blocks mapped to devices
per iteration~\cite{jiang2025dcp}, and MagiAttention balances
mask-dependent attention work across context-parallel ranks with a
dispatch solver~\cite{magiattention2025}; both redistribute attention
computation, whereas \textsc{Layerwise Repacking} changes where a sample's
persistent sparse-trunk state lives.

\subsection{Sparse Attention Parallelism}
Sparse Transformer established structured attention sparsity as a route to
subquadratic long-sequence modeling~\cite{child2019sparse}.  db-SP balances
sparse-attention workload at the head and block levels while selecting
parallel degrees dynamically for visual-generation inference~\cite{chen2026dbsp}.
Native Sparse Attention combines token compression with blockwise token
selection to make sparse attention trainable and hardware
aligned~\cite{yuan2025nsa}; such mechanisms bound or compress each query's
attended keys and so shrink the attention term, leaving the sparse layers
profiled here dominated by near-linear projection and MoE work.
Our training setting instead targets a layer-level mismatch between wide full
attention and node-local sparse execution.  The sparse layers profiled here
are MoE-dominated; expert-parallel systems balance token routing itself
through dispatch/combine all-to-all~\cite{lepikhin2021gshard}, whereas our
planner treats per-token expert work as approximately length-linear cost when
placing samples and leaves the expert-routing code unchanged.

\subsection{Layer-Wise Parallelism Planning}
Automatic parallelism systems already choose strategies per layer or per
stage: Galvatron selects a hybrid of data, sharded-data, tensor, and
pipeline parallelism for each Transformer layer by dynamic
programming~\cite{miao2022galvatron}, and Alpa jointly plans inter- and
intra-operator parallelism~\cite{zheng2022alpa}.  MoE Parallel Folding
decouples the parallel mappings of attention and MoE layers so that each
layer type adopts its own configuration~\cite{liu2025moefolding}.  These
mappings are fixed for a model and cluster before training.  HotSPa varies
the strategy within a step, applying different parallelism strategies to
different groups of sequences and transferring parameters and gradients
when it switches~\cite{ge2024hotspa}; each sequence group keeps one
strategy across all layers.  \textsc{Layerwise Repacking} instead changes, in every
step, which ranks hold each sample as execution moves between full and
sparse layers, driven by that step's sample mix, and makes the change
inside the mandatory Ulysses exchange.

\subsection{Variable-Length Workload Placement}
FlexSP assigns each sequence a sequence-parallel degree by solving a linear
program over profiled costs~\cite{wang2025flexsp}; ByteScale unifies data
and context parallelism with a dynamic communication mesh and a balance
scheduler~\cite{ge2025bytescale}; and WLB-LLM balances workload with
variable-length document packing and per-document context-parallel
sharding~\cite{wang2025wlbllm}; and Megatron-Core's Dynamic Context
Parallelism, which our baseline follows, selects a context-parallel size per
packed micro-batch~\cite{li2026dynamiccp}.  NSP nests SP groups of
different sizes on shared GPUs within one iteration, so long sequences use
larger groups and short ones smaller groups~\cite{wang2026nsp}.  In these
placement systems one group serves every layer of a sample; Layerwise
Repacking instead changes a sample's group between layer types within one
step.  KnapFormer combines global knapsack assignment with sequence
parallelism for mixed-resolution DiT training and redistributes tokens with
all-to-all at the two ends of the transformer trunk~\cite{zhang2025knapformer};
its groups then also serve every layer.  The per-boundary dedicated pass
that Table~\ref{tab:dedicated-pass} estimates is our own alternative
design, not a measurement of KnapFormer.  Zellige organizes mixed image/video
placement around heavy anchors and light fillers~\cite{xiang2026zellige}.
The candidate generators of our improved solver draw on greedy
list scheduling~\cite{graham1969bounds} and large/small-job
decomposition~\cite{hochbaum1987dual}; what distinguishes the solver is its
two-component cost model and evaluator-based selection across candidates,
while the runtime performs the selected repacking through the mandatory
Ulysses exchange.

\section{Conclusion}

Variable-length hybrid-attention training creates distinct parallelism
requirements for full and sparse layers, but adopting layer-specific
layouts changes token ownership. \system{} retains wide parent groups
for full attention and repacks samples into node-local children for
sparse layers. Fusing layout conversion into the two mandatory
full-layer Ulysses exchanges eliminates cross-node sparse-layer Ulysses
communication for repacked parents without adding a separate
redistribution pass. An auxiliary parent solver reassigns samples
across fixed parent groups to improve the load balance achievable
by the children.

Across all 22 steps containing SP16/32 parents in the first 200 training
steps, repacking alone reduces E2E by 5.95\% and 4.17\% under the baseline
and improved solvers, and the complete system reduces E2E by 7.61\% and
raises MFU by 4.83\pp{} over the baseline with repacking disabled.  The
gains hold on all 20 one-SP16 workloads; on the two-SP16 and SP32
workloads, the SP8 child cap leaves one long sample dominating its parent,
and the resulting imbalance erases the gain.  Gating repacking by predicted
gain, making the child cap a per-sample choice, scoring the executed plan,
and evaluating more multi-parent and SP32 steps are the next steps.

\section*{Open-Source Software Statement}

This work uses the open-source project TorchTitan, released under the
BSD 3-Clause License; our training system and the \system{} implementation
are built on it.  We strictly adhered to its license requirements; the
original project's copyright notice and license text can be found in its
official repository at \url{https://github.com/pytorch/torchtitan}.

\bibliographystyle{ACM-Reference-Format}
\bibliography{references}

\appendix

\section{Solver Route Constructions}
\label{app:solver-routes}

\begin{algorithm}[t]
\caption{Resource-Priority Parent Allocation}
\label{alg:resource-lpt}
\small
\begin{algorithmic}[1]
\Require Samples $\mathcal I$, fixed parents $\mathcal B$, keep set
         $\mathcal K$, and priority $\nu\in\{\mathrm{compute},\mathrm{memory}\}$
        ;
\Ensure Valid parent assignment $F$, or failure;
\State Initialize $I_p\gets\emptyset$ and
       $\ell_p\gets\ell_0(R_p)$ by Equation~\eqref{eq:initial-parent-load};

\Statex
\Statex \textbf{Anchor Placement}
\For{each $i\in\mathcal K$ in descending $(a_i,-i)$ order}
    \State Restrict candidates to distinct empty SP$>8$ parents;
    \State Select $p_i^\star$ by Equation~\eqref{eq:lpt-placement}
           under this restriction;
    \If{no candidate survives}
        \State \Return failure;
    \EndIf
    \State $I_{p_i^\star}\gets I_{p_i^\star}\cup\{i\}$ and update
           $\ell_{p_i^\star}$ by Equation~\eqref{eq:sample-marginal};
\EndFor

\Statex
\Statex \textbf{Priority Allocation}
\State Sort $U\gets\mathcal I\setminus\mathcal K$ by
       $Q_i^\nu$ in Equation~\eqref{eq:priority-key};
\For{each $i\in U$ in descending priority order}
    \If{remaining samples equal empty parents}
        \State Restrict $\mathcal B_R^{(t)}$ to empty parents;
    \EndIf
    \State Construct $\mathcal R_i$ and select $p_i^\star$
           by Equation~\eqref{eq:lpt-placement};
    \If{$\mathcal R_i=\emptyset$}
        \State \Return failure;
    \EndIf
    \State $I_{p_i^\star}\gets I_{p_i^\star}\cup\{i\}$;
    \State $\ell_{p_i^\star}\gets
           \ell_{p_i^\star}+m_i(R_{p_i^\star})$;
\EndFor

\Statex
\Statex \textbf{Validation and Scoring}
\State Verify coverage, non-empty parents, anchors, and length caps;
\If{any invariant fails}
    \State \Return failure;
\EndIf
\State Rebuild $F$ and compute $\widehat T(F)$ by
       Equation~\eqref{eq:parent-eval};
\State \Return $F$
\end{algorithmic}
\end{algorithm}

\begin{algorithm}[t]
\caption{Anchor-Fill Parent Allocation}
\label{alg:anchor-fill}
\small
\begin{algorithmic}[1]
\Require Samples $\mathcal I$, fixed parents $\mathcal B$, and keep set
         $\mathcal K$;
\Ensure Valid parent assignment $F$, or failure;
\State Initialize $(I_p,\ell_p)$ by Equation~\eqref{eq:initial-parent-load};
       disable the length cap;
\State Order $U\gets\mathcal I\setminus\mathcal K$ ascending by $(a_i,i)$;
\For{each $i\in\mathcal K$ in descending $(a_i,-i)$ order}
    \State Choose a distinct empty SP$>8$ parent by
           Equation~\eqref{eq:lpt-placement}; fail if none exists;
    \State Assign $i$ and update its parent by
           Equation~\eqref{eq:sample-marginal};
\EndFor
\State Let $\mathcal P_A$ follow anchor order; reserve one sample for each
       unanchored empty parent;
\State Compute $\tau$ and $\beta_p$ by Equation~\eqref{eq:anchor-budget};
\For{each $p\in\mathcal P_A$ in anchor-placement order}
\State $A_p\gets\langle\rangle$ and $h_p\gets\sum_{j\in I_p}a_j$;
    \While{$U\ne\emptyset$ and the empty-parent reserve is preserved}
        \State Select the cheapest feasible $f_p^\star$ by
               Equation~\eqref{eq:anchor-budget};
        \If{$f_p^\star$ exists}
            \State Move $f_p^\star$ from $U$ to $A_p$ and update $h_p$;
        \Else
            \State If $A_p\ne\emptyset$, swap
                   $\operatorname{first}(A_p)$ with
                   $\operatorname{first}(U)$ and allow one budget overflow;
            \State \textbf{break}
        \EndIf
    \EndWhile
    \State Commit $A_p$ to $p$, update $\ell_p$ per filler by
           Equation~\eqref{eq:sample-marginal}, and close $p$;
\EndFor
\State Order residual $U$ by $Q_i^{\mathrm{compute}}$ and allocate it by
       Equations~\eqref{eq:priority-key}--\eqref{eq:lpt-placement};
\State Verify coverage, non-empty parents, and anchors; fail on violation;
\State Rebuild $F$ and compute $\widehat T(F)$ by
       Equation~\eqref{eq:parent-eval};
\State \Return $F$
\end{algorithmic}
\end{algorithm}

This appendix specifies the route constructions and the selection rule
summarized in Section~\ref{sec:planner}.

Let $\mathcal K\subseteq\mathcal I$ denote the keep set: heavy samples
whose predicted work exceeds a fixed multiple of the mean per-rank work,
each of which must anchor a distinct wide parent (the solve falls back to
the round-1A membership if $|\mathcal K|$ exceeds the number of wide parents).
$\mathrm{cap}(i,p)$ holds if assigning sample $i$ to parent $p$ satisfies
the baseline per-parent length cap; Anchor-Fill does not apply it.
$\epsilon(x,y)$ is the fixed numerical tolerance used only for
deterministic tie handling.

\subsection{Resource-Priority Allocation}

Both resource-priority routes are longest-processing-time (LPT) style
greedy allocators that differ only in their priority key.
Let $m_i^{\max}=\max_R m_i(R)$ and $m_i^{\min}=\min_R m_i(R)$, taken over
the SP classes present in the fixed topology.  The descending lexicographic
queue keys are
\begin{equation}
\begin{aligned}
  Q_i^{\mathrm{compute}}
    &=(m_i^{\max},m_i^{\min},L_i,-i),\\
  Q_i^{\mathrm{memory}}
    &=(L_i,m_i^{\max},-i).
\end{aligned}
\label{eq:priority-key}
\end{equation}
$L_i$ serves as a token-count resource proxy.  Each route sorts samples by one
key in descending order.  Required heavy samples are pre-placed; for every remaining
sample, the allocator asks each SP class for its least-loaded feasible parent
and chooses among those representatives by projected global makespan.
Compute-LPT therefore gives expensive samples first access to scarce wide
parents.  Memory-LPT gives long samples first access to length-feasible
capacity, reducing fragmentation by short samples.  The allocation skeleton,
feasibility predicates, and deterministic tie breaks are otherwise identical.

For a placement step $t$, let $\mathcal B_R^{(t)}$ contain only empty SP-$R$
parents when the number of remaining samples equals the number of empty
parents, and all SP-$R$ parents otherwise.  The route takes the
least-loaded length-feasible parent of each SP class as that class's
representative and places the sample on the representative that minimizes
the projected global makespan:
\begin{equation}
\begin{aligned}
  \mathcal R_i
    &=\Bigl\{\operatorname*{arg\,min}_{
        p\in\mathcal B_R^{(t)}:\,\mathrm{cap}(i,p)}\ell_p\Bigr\}_{R},\\
  p_i^\star
    &=\operatorname*{arg\,min}_{p\in\mathcal R_i}
       \max\!\Bigl(\max_{b\ne p}\ell_b,\ \ell_p+m_i(R_p)\Bigr).
\end{aligned}
\label{eq:lpt-placement}
\end{equation}
Here $\mathcal R_i$ contains the least-loaded feasible
parent from each eligible SP class, and $p_i^\star$ is the selected
parent for sample $i$.
Near-ties within the tolerance $\epsilon$ prefer the smaller projected
parent load, then the smaller SP class, then a fixed rank order, so
identical inputs produce identical assignments.

Algorithm~\ref{alg:resource-lpt} summarizes the shared allocation skeleton for
the compute- and memory-priority views.

\subsection{Anchor-Fill Allocation}

Anchor-Fill starts from the structural decision.  It places each
threshold-selected keep sample in a distinct eligible large parent; only these
anchored parents enter the fill loop, so the number of anchors may be smaller
than the number of large parents.  It accepts cheap fillers while they fit
below a fixed multiple of the global mean predicted work.  When the next cheapest filler does not
fit, it may replace the cheapest accepted filler once, intentionally allowing
the parent to exceed that budget, then closes the parent and sends the remainder
through the shared compute-LPT constructor.  This creates a qualitatively different
candidate: large parents are not merely containers for the most expensive
samples; they also absorb flexible work that their full-layer parallelism can
carry before round~2 redistributes that parent locally.

For an anchored parent $p$, let $h_p$ be its current predicted
work and let
$U$ be the unassigned pool.  Anchor-fill scales the global mean per-rank work by a fixed factor
$\kappa\ge1$ into a parent-sized budget, then chooses the cheapest filler
that still fits:
\begin{equation}
\begin{aligned}
  h_p &= \sum_{j\in I_p}a_j,
  \tau   = \frac{\sum_i a_i}{\sum_b R_b},
  \beta_p = \kappa\tau R_p,\\
  \mathcal U_p
    &=\{i\in U\mid h_p+a_i\le \beta_p\},\\
  f_p^\star
    &=\operatorname*{arg\,min}^{\mathrm{lex}}_{i\in\mathcal U_p}(a_i,i).
\end{aligned}
\label{eq:anchor-budget}
\end{equation}
Here $\tau$ is the mean reference work per rank,
$\beta_p$ is parent $p$'s work budget, and $\mathcal U_p$ is its set
of feasible fillers; $\mathrm{lex}$ denotes lexicographic selection
by $(a_i,i)$.

Unlike Zellige, which places anchors with CP-SAT~\cite{xiang2026zellige},
the route preserves the round-1A parent count, sizes, and rank intervals;
uses a greedy heavy assignment and one optional swap; and later allows
round~2 to split the resulting parent into sparse children.

Algorithm~\ref{alg:anchor-fill} summarizes the anchor-fill candidate.
In Algorithm~\ref{alg:anchor-fill}, $\mathcal P_A$
lists the anchored parents and $A_p$ lists the fillers accepted by
parent $p$, both in placement order; $\langle\rangle$ is an empty
list and $\operatorname{first}(\cdot)$ selects its first element.

\subsection{Candidate Selection Rule}

Let $\mathcal V$ be the set of enabled routes that return a valid candidate
$F_r$.  Final selection uses predicted E2E with a numerical tolerance and the fixed route order Compute-LPT,
Memory-LPT, Anchor-Fill:
\begin{equation}
\begin{aligned}
  \widehat T_{\min}&=\min_{r\in\mathcal V}\widehat T(F_r),\\
  \mathcal A
    &=\{r\in\mathcal V\mid
      \widehat T(F_r)\le \widehat T_{\min}
      +\epsilon(\widehat T(F_r),\widehat T_{\min})\},\\
  r^\star
    &=\operatorname*{arg\,min}_{r\in\mathcal A}\pi_{\mathrm{route}}(r),
  \quad
  \pi_{\mathrm{route}}=(0,1,2).
\end{aligned}
\label{eq:route-selection}
\end{equation}
Here $\widehat T_{\min}$ is the minimum predicted time,
$\mathcal A$ contains the routes tied within tolerance, and $r^\star$
is the selected route. The function $\pi_{\mathrm{route}}$ assigns
priorities $0$, $1$, and $2$ to \textsc{Compute-LPT},
\textsc{Memory-LPT}, and \textsc{Anchor-Fill}, respectively.

\end{document}